\documentclass[
twocolumn,
]{ceurart}

\usepackage[frozencache,cachedir=.]{minted}
\setminted{breaklines=true}

\usepackage{xcolor}
\usepackage{amsmath}
\usepackage{amssymb}
\usepackage{multirow}
\usepackage{array}

\newcommand{\bos}{\textit{$<$bos$>$}}

\begin{document}

\copyrightyear{2021}
\copyrightclause{Copyright for this paper by its authors.
  Use permitted under Creative Commons License Attribution 4.0
  International (CC BY 4.0).}

\conference{CLiC-it 2023: Ninth Italian Conference on Computational Linguistics,
  November 30--December 2, 2023, Venice, IT}

\title{How To Build Competitive Multi-gender Speech Translation Models For Controlling Speaker Gender Translation}

\author[1]{Marco Gaido}[%
email=mgaido@fbk.eu,
]
\address[1]{Fondazione Bruno Kessler}

\author[1,2]{Dennis Fucci}[%
email=dfucci@fbk.eu,
]
\address[2]{University of Trento}

\author[1]{Matteo Negri}[%
email=negri@fbk.eu,
]

\author[1]{Luisa Bentivogli}[%
email=bentivo@fbk.eu,
]

\begin{abstract}
When translating from 
%gender-neutral
notional gender
languages (e.g., English) into grammatical gender languages (e.g., Italian), the generated translation requires explicit gender assignments for various words, including those referring to the speaker. When the source sentence does not convey the speaker's gender,
speech translation (ST) models either rely on the possibly-misleading vocal traits of the speaker or default to the masculine gender, the most frequent in existing training corpora. To avoid such biased and not inclusive behaviors, the gender assignment of speaker-related expressions should be guided by externally-provided metadata about the speaker's gender.\footnote{\label{foot:gender_disclaimer}Throughout the paper, we use the word \textit{gender} to indicate the preferred linguistic expression of gender and not the gender identity.} While previous work has shown that the most effective solution is represented by separate, dedicated gender-specific models, the goal of this paper is to achieve the same results by integrating the speaker's gender metadata into a single ``multi-gender'' neural ST model, easier to maintain. Our experiments demonstrate that a single multi-gender model outperforms gender-specialized ones when trained from scratch (with gender accuracy gains up to 12.9 for feminine forms), while fine-tuning from existing ST models does not lead to competitive results.

\end{abstract}


\maketitle

\section{Introduction}
\setcounter{footnote}{1}
\footnotetext{Throughout the paper, we use the word \textit{gender} to indicate the preferred linguistic expression of gender and not the gender identity.}\label{foot:gender_disclaimer}

Spurred by growing concerns about fairness in language technologies, research on understanding and mitigating gender bias in automatic translation is gaining traction \cite{savoldi-etal-2021-gender}.
The bias of automatic systems is extremely evident when it comes to ambiguous sentences or expressions, where there are no explicit cues in the source content about the correct gender\textsuperscript{\ref{foot:gender_disclaimer}} assignment of a 
%referent. 
referent (e.g., en: \textit{The doctor arrived} -- it: \textit{\underline{Il}/\underline{La} dottor\underline{e}/\underline{essa} è arrivat\underline{o}/\underline{a}}).
In this setting, the state-of-the-art neural models often choose the masculine forms or perpetuate stereotypical assignments, as they reflect the condition statistically more likely 
% according to 
based on
their (biased) training data \cite{Prates2018AssessingGB,cho-etal-2019measuring}.

This situation frequently occurs when the source language is genderless or employs notional gender,
expressing gender in a limited set of parts of speech, and the target language follows a grammatical gender system, embedding gender distinctions throughout a broad inventory of parts of speech.
%% OLD
% This situation often arises 
% \df{when \mg{the source language % conveys gender}
% \mn{only through a few gender-% specific nouns (e.g., % \textit{mother}, \textit{brother}, % \textit{wife}) and\mg{/or} pronouns % (e.g., \textit{she}, \textit{her}, % \textit{he}, \textit{him}), as in % genderless or notional gender % languages, 
% \mg{and the target language embeds % gender} into a large inventory of % parts of speech \mn{(i.e. nouns, % adjectives, determiners, verbs, % pronouns)}, as in grammatical % gender languages.}}
%
Focusing on the case in which the source language is 
% English (a notional gender language) 
English, a notional gender language,
and the target language is 
% Italian (a grammatical gender language), 
Italian, a grammatical gender language,
a frequent instance of this condition is represented by first-person references, 
i.e. by the words and expressions referred to the speaker (e.g., en: \textit{I am a young researcher} -- it: \textit{Sono \underline{una}/\underline{un} giovane ricerca\underline{trice}/\underline{tore}}).
In this case, text-to-text machine translation (MT) models mostly output masculine forms, while 
direct (or end-to-end) speech-to-text translation (ST) systems partly rely on the biological cue of the speaker's vocal traits to assign gender \cite{bentivogli-etal-2020-gender,gaido-2020-on-knowledge}.
However, direct ST models are still largely biased toward producing masculine forms, and, most importantly, biological aspects are related to the sex rather than to the gender of an individual.
Hence,
their exploitation is not inclusive of all people, harming several groups such as transgenders \cite{zimmantransgender}.

As a solution, \cite{gaido-etal-2020-breeding} proposed to leverage external metadata about the speaker's gender to control the gender assignment of words referred to the speaker.
Specifically, they investigated two approaches: \textit{i)} the development of two separate \textit{gender-specialized models}, fine-tuned on gender-specific data as also proposed later in MT \cite{choubey-etal-2021-gfst}, and \textit{ii)} a single \textit{multi-gender model}, where the speaker gender is a tag fed to 
%the model
a single model
as in multilingual systems \cite{johnson-etal-2017-googles}.
%\cite{ha-etal-2016-toward,johnson-etal-2017-googles,Inaguma-2019-multilang,di-gangi-2019-multilingual}.
While the second solution would be preferable (as the specialized solution involves the higher cost of maintaining two separate models), the experiments in \cite{gaido-etal-2020-breeding} demonstrate that specialized models outperform the multi-gender approach by a large margin in terms of gender accuracy.
% This is also true in a counterfactual scenario, namely when vocal characteristics are opposed to the typical gender they are associated to.

% In this paper, we reconsider the multi-gender approach to understand why it does not achieve competitive results compared to specialized systems. 
In light of the above, in this paper we address the following research questions: \textit{i)} why do specialized models outperform multi-gender ones? \textit{ii)} Can we build competitive multi-gender systems?
% We conduct experiments on English-Italian TED talk translation and find that the low accuracy of multi-gender models is due to their initialization with gender-unaware ST system weights and the inability to override the reliance on vocal cues during fine-tuning. Training a multi-gender model from scratch outperforms the specialized approach.
Through experiments on English-Italian translation of TED talks, we show that the low accuracy of multi-gender models comes from the initialization with the weights of a gender-unaware ST system and the inability to override the behavior of the base ST model (i.e., the reliance on vocal cues) during the fine-tuning stage. 
We also try to address this problem with two solutions: \textit{i)} a contrastive loss that penalizes the extraction of gender cues from speech input, 
and \textit{ii)} altering vocal properties of training data to misalign gender cues with gender tags and gender translations.\footnote{Our code is released open source under Apache 2.0 Licence at: \url{https://github.com/hlt-mt/FBK-fairseq/}} 
% While these solutions bring some benefits in gender accuracy and overall translation quality compared to a multi-gender model fine-tuned from a base model, they still do not
Despite the slight improvements brought by these solutions in gender accuracy and overall translation quality, 
none of them effectively close the performance gap with the specialized solution.
%None of the two solutions, though, effectively bridges the performance gap with the specialized solution.
However, training multi-gender models from scratch yields competitive results, outperforming the specialized approach with gender accuracy gains of up to 12.9 points for feminine translations.
%Notably, these benefits are not available in a multi-gender model trained from scratch.
%
%As such, we recommend building multi-gender models from scratch, while building them on top of existing systems remains an open research question.
Therefore, we recommend building multi-gender models from scratch, while building them on top of existing systems remains an open research question.

\section{Background}

In this section, we introduce the basic concepts useful for understanding the rest of the paper. First, we provide an overview of the methods proposed in the literature to integrate language tags into neural multilingual translation models (\S\ref{sec:back_multiling}), from which multi-gender models draw inspiration. Then, we present how gender information has been removed from neural representations through adversarial training in previous works (\S\ref{sec:back_gender_removal}), from which we derive our solution presented in \S\ref{sec:gradient_reversal}.
%.
% \df{, %.
% We adopt this strategy to penalize the model from relying on acoustic information related to % gender.}

\subsection{Tags Integration in Multilingual Models}
\label{sec:back_multiling}

State-of-the-art models in MT and ST are sequence-to-sequence models made of an encoder and an autoregressive Transformer decoder \cite{transformer}.
The autoregressive decoder predicts the next-token probability over a predefined vocabulary at every iteration by looking at the encoder output and at the previously generated tokens, which are pre-pended a special token named \textit{beginning of sentence} (\bos). Formally, the probability $p_V(y_t)$ over the vocabulary $V$ at time step $t$ is:
\begin{equation}
\label{eq:decoding}
\text{softmax}(D(E(X); \text{\bos}, y_0, ..., y_{t-1}))
\end{equation}
where \textit{E} is the encoder, \textit{D} is the decoder, \textit{X} is the input sequence, and $y_i$ is the token generated at the $i$-th time step.

While early attempts to build multilingual MT models were based on training dedicated encoders and decoders for each language \cite{firat-etal-2016-multi,firat-etal-2016-zero}, nowadays the preferred solution is a model made of a single universal encoder and decoder where the language is represented as a tag pre-pended to the text \cite{sennrich-etal-2016-improving,ha-etal-2016-toward,johnson-etal-2017-googles}. In the case of one-to-many multilingual models, this means that the \bos{} token is replaced with a token that indicates the language, so that Eq. \ref{eq:decoding} becomes:
\begin{equation}
\label{eq:decoding2}
\text{softmax}(D(E(X); \text{LID}, y_0, ..., y_{t-1}))
\end{equation}
where \textit{LID} is the identifier of the desired target language.

In direct ST, \cite{Inaguma-2019-multilang} demonstrated the effectiveness of this solution, also known as ``target forcing'', while \cite{di-gangi-2019-multilingual} proposed other methods to integrate the language information into the architecture.
Thanks to its simplicity and effectiveness, target forcing is currently the most widespread method to build multilingual ST systems \cite{wang2020fairseqs2t,salesky21_interspeech}, also when using large pre-trained textual models such as mBART \cite{liu-etal-2020-multilingual-denoising} to initialize the ST decoder \cite{liu-etal-2023-kits,gow-smith-etal-2023-naver}.
%, thanks to its simplicity and effectiveness. 
In line with this trend, \cite{gaido-etal-2020-breeding} obtained their best multi-gender models with target forcing. As such, we build multi-gender models using target forcing with the \textit{F} and \textit{M} tags representing the two grammatical genders instead of the language identifiers.

\subsection{Gender Information Removal}
\label{sec:back_gender_removal}

With the goal of fairer technology that does not rely on spurious cues reflecting
%, which reflect 
stereotypical biases 
%reflected 
in the available data, researchers have tried to build systems that achieve ``equalized odds'' among different demographic groups \cite{10.5555/3157382.3157469}. Formally, this means that, given an attribute $z$ representing the belonging to one of the $Z$ demographic groups, the predicted probability $p(\hat{Y})$ of a fair system should be independent of the variable $z$, i.e. $p(\hat{Y}) = p(\hat{Y} | z), \forall z\in Z$. The variable $z$ is named the \textit{protected attribute} and in the context of gender bias literature represents the gender of the involved person. So in this work we consider $Z=\{F, M\}$.\footnote{Although this paper does not aim at perpetuating a binary vision of gender, in this work we limit to the feminine and masculine categories for the sake of simplicity, as the available benchmarks currently do not cover the non-binary case.}

The first attempts to achieve equalized odds across genders in neural systems have focused on deep neural network (DNN) classifiers \cite{46295,10.1145/3278721.3278779,elazar-goldberg-2018-adversarial,PPR:PPR540557}. In this line of work, the last hidden representation of the DNN is passed both to a linear layer that predicts the classification scores $\hat{Y}$ and to a linear layer (the \textit{discriminator}) devoted to predicting the protected attribute $z$. The DNN is then trained in an adversarial manner \cite{10.1145/3422622}, i.e. it is alternatively trained to \textit{i)} predict $z$ (while keeping the shared DNN freezed) and \textit{ii)} predict $\hat{Y}$ while minimizing the ability to predict $z$ (keeping the protected attribute classification layer freezer). 
As this training procedure is often unstable, similar practices based on minmax optimizations have been proposed \cite{pmlr-v162-ravfogel22a}, even with discriminators made of functions different from linear projections \cite{ravfogel-etal-2022-adversarial}, or using more than one discriminator \cite{han-etal-2021-diverse}. \cite{shao-etal-2023-erasure} also proposed methods to automatically extract the protected attributes in case they have not been provided.

Such adversarial training procedures can be seen as an extension of the \textit{gradient reversal} \cite{pmlr-v37-ganin15}, where the training alternatively freezes the base model (to refine the discriminator) and the discriminator (inverting its loss to train the base model to be unable to discriminate). In fact, the gradient reversal layer, applied to the hidden representations before feeding them to the discriminator, is an identity function in the forward pass, while inverts the gradient in the backward pass, scaling it by a positive factor $\lambda$. By naming $x$ the hidden representation, $D$ the discriminator, and $GRL$ the gradient reversal layer, this means that:
\begin{equation}
\label{eq:gradient_reversal}
GRL(x) = x, \nabla D(GRL(x)) = -\lambda \nabla D(x)
\end{equation}
where $\lambda$ is a hyperparameter that can either be fixed or updated according to the following equation:
\begin{equation}
\label{eq:lambda}
\lambda = \frac{2}{1+e^{-\gamma p}} - 1
\end{equation}
where $p$ is the ratio between the number of parameter updates performed and the total number of updates needed to complete the training.

\section{Solutions for Multi-gender Models}

%\dfff{To investigate and solve the potential factors that impact the performance of multi-gender models, we explored several training techniques. 
%Firstly, we replicated the multi-gender models as described in \cite{gaido-etal-2020-breeding}, namely by fine-tuning from a base ST system.
%Secondly, we trained multi-gender models from scratch, using the gender tag since the beginning of the training.
%This was done to examine whether the issue with the multi-gender technique could be related to weight initialization with a base system.
%In addition, we implemented two further approaches to create multi-gender ST models that rely solely on the gender tag, disregarding any spurious cues related to speakers' vocal traits.
%In \S\ref{sec:gradient_reversal} we discuss our attempt to establish gender-invariant encoder representations by incorporating a gradient-reverted discriminator on the speaker's gender.}
To create multi-gender ST models that solely rely on the gender tag, ignoring spurious cues related to speakers' vocal traits, we test two approaches.
First, we try to create gender-invariant encoder representations by adding a gradient-reverted discriminator on the speaker's gender (\S\ref{sec:gradient_reversal}). 
Second, we manipulate the input audio by altering the speaker's pitch, so that the correlation between the gender tag (and output text) and the speaker's vocal traits is lost (\S\ref{sec:audio_manip}).
%\dfff{In \S\ref{sec:audio_manip} we 
%we explain our strategy to break the correlation between the gender tag (and the target gender translation) and the speaker's vocal traits by manipulating the speaker's vocal characteristics. 
%We experimented with these techniques using both fine-tuned and trained from scratch models.}

\subsection{Gradient Reversal}
\label{sec:gradient_reversal}

As seen in \S\ref{sec:back_multiling}, the decoder of a multi-gender model has three inputs: the encoder output, a tag representing the speaker's gender, and the previously generated tokens. As we want the decoder to have the tag as the only source of information about the speaker's gender, we propose to create encoder outputs that do not convey any information regarding the speaker's gender by adding a gradient-reverted discriminator on top of the encoder, motivated by the success of this approach in MT with sentences where there is a single referent whose gender has to be determined \cite{fleisig2022mitigating}. The discriminator is made of two fully-connected layers with ReLU activation function \cite{10.5555/3104322.3104425}, whose output is averaged over the temporal dimension to obtain a single vector representing the logit\footnote{The \textit{logit} is the vector of raw predictions before a function (commonly, the softmax) that maps it into probabilities.} of the discriminator.

Furthermore, we experiment with assigning dedicated class weights to the loss of the discriminator, as a countermeasure to the class imbalance between female and male speakers in the training data. Specifically, we assigned the weights ($w_f, w_m$) proportionally to the inverse of the frequency of each class ($f_f, f_m$) in the training data:

\begin{equation}
\nonumber
\label{eq:ww}
\begin{cases}
w_f \propto \frac{1}{f_f}, w_m \propto \frac{1}{f_m} \\
w_f * \frac{f_f}{f_m + f_f} + w_m * \frac{f_m}{f_m + f_f} = 1
\end{cases}
\end{equation}

resulting in $w_f=1.4$, $w_m=0.8$ in our case.

\subsection{Audio Manipulation}
\label{sec:audio_manip}

Our second approach aims to break the correlation between the vocal characteristics of the speaker on one side and the gender tag and target translation on the other.
To this aim we manipulate part of the training data using the \textit{Opposite} pitch manipulation strategy by \cite{fucci2023pitch}. The amount of data that is manipulated at each iteration (epoch) is controlled with a hyperparameter, $p$, which determines the probability of altering an utterance, regardless of whether it is produced by a male or female speaker.
%that we keep equal for male and female speakers. 
The manipulation is performed by altering two crucial acoustic parameters distinguishing between male and female voices \cite{Coleman,Hillenbrand}: $f0$ and formants.
% Our second approach aims to break the correlation between the vocal characteristics of the speaker
% %\df{on one side}
% and the gender tag and target 
% %\df{gender} 
% translation %. 
% %\df{on the other.
% by manipulating the audio of part of the training data with the \textit{Random} pitch manipulation strategy by \cite{fucci2023pitch}. 
% %We achieve this by manipulating the audio of a portion of the training data.}
% The amount of data that is manipulated at each iteration (epoch) is controlled with a hyperparameter, $p$, which determines the probability of altering an utterance.
% Each voice can be manipulated to exhibit the characteristics of either the same or the opposite gender (the choice is random), and the manipulation
% % The conversion of a female/male voice (toward a male/female voice) 
% is performed by manipulating two crucial acoustic parameters distinguishing between male and female voices \cite{Coleman,Hillenbrand}: $f0$ and formants.
% % In the given audio segment uttered by a speaker, we align them with values typical of the opposite gender.
% % This method allows us to obtain some (manipulated) training data which exhibit acoustic characteristics not aligned with the gender tag.
% % Using
In particular, we first estimate the $\tilde{f\mathit{0}}$ median of the $f0$ contour of the considered speech segment. Then, we sample a new $\tilde{f\mathit{0}}'$ median of the desired output audio from a 
normal distribution whose mean and standard deviation depend on the target gender: 
for feminine voices, we use 250 Hz as the mean and 17 as the standard deviation so that the sampled value is between 199 Hz and 301 Hz with 99.7\% probability; for masculine voices, the mean is 140 Hz and the standard deviation is 20 to obtain a 99.7\% probability range within 80 Hz and 200 Hz.
Once $\tilde{f\mathit{0}}$ and $\tilde{f\mathit{0}}'$ are defined, we compute a scaling factor $\alpha$ as the ratio $\tilde{f\mathit{0}'} / \tilde{f\mathit{0}}$.
Lastly, 
%\df{Subsequently,} 
the original $f0$ contour is scaled by the $\alpha$ factor, 
%and 
while
the formants are scaled by 1.2 when converting from male to female voices, or by 0.8 otherwise.
%we independently shift $f0$ and formants.
%For $f0$, we define a value representing the $f0$ median ($\tilde{f\mathit{0}}'$) of the new %manipulated audio.
%$\tilde{f\mathit{0}}'$ is sampled from a normal distribution whose mean and standard deviation %depend on the target gender defined by the chosen policy: 
%A scaling factor is then computed as the ratio $\tilde{f\mathit{0}'} / \tilde{f\mathit{0}}$, 
%which scales the pitch contour to achieve the new target contour.
%Regarding formants, we set the scaling factor to 1.2 if the target gender is feminine, 0.8 if %it is masculine.
%We also control the amount of segments to be manipulated by means of a probability value $p$, %which determines the probability of manipulating a segment uttered by a female voice (toward a %male voice) or by a male voice (toward a female voice).
This perturbation is applied independently to each sample during each training epoch, so as to maximize the variability of the training data.
% ,
% and the scaling operations are performed using
% the ``Change gender'' function\footnote{\url{https://www.fon.hum.uva.nl/praat/manual/Sound__Change_gender___.html.}} of Praat \cite{praat}.

\begin{table*}[t]
\small
\centering
\setlength{\tabcolsep}{10pt}
\begin{tabular}{l|c|>{\centering}m{1.2cm} >{\centering}m{1.2cm}||cc}
\multirow{2}{*}{\textbf{Model}} & \multirow{2}{*}{\textbf{BLEU ($\uparrow$)}} & \multicolumn{4}{c}{\textbf{Gender Accuracy ($\uparrow$)}} \\
\cline{3-6}
 & & \multicolumn{1}{c}{\textbf{1F}} & \multicolumn{1}{c||}{\textbf{1M}} & \multicolumn{1}{c}{\textbf{1F-Tag M}} & \multicolumn{1}{c}{\textbf{1M-Tag F}} \\
\hline
\multicolumn{6}{c}{\textit{Fine-tuning}} \\
\hline
Specialized  & \textbf{27.4}  & 73.3 & 92.5  & 80.9  & 56.1 \\
\hline

Multi-gender   & 26.0  & 66.8  & 78.0  & 64.6  & 47.1  \\
+ gradient reversal  & 26.7  & 60.7  & 85.9  & 77.0  & 43.3  \\
+ gradient reversal weighted & 26.3  & 62.4  & 83.5  & 77.7  & 45.9  \\
+ audio manipulation (50\%) & 26.4  & 56.0  & 82.6  & 60.7  & 33.9  \\
+ audio manipulation (80\%) & 26.3  & 69.3  & 81.1  & 69.0  & 47.7 \\

\hline
\multicolumn{6}{c}{\textit{Training from scratch}} \\
\hline
Multi-gender  & 27.2  & \textbf{84.0}  & 92.7  & 93.4  & \textbf{69.0}  \\
+ gradient reversal & 24.9  & 70.9  & \textbf{93.2}  & \textbf{94.1}  & 58.7  \\
+ gradient reversal weighted & 24.2  & 75.8  & 92.6  & 92.8  & 63.5  \\
+ audio manipulation (50\%) & 26.2  & 79.6  & 92.4  & 91.5  & 67.9  \\
+ audio manipulation (80\%) & 25.7  & 81.7  & 92.6  & 91.0  & 65.3  \\

\end{tabular}
\caption{BLEU and gender accuracy scores for the specialized models (Specialized) and the multi-gender models (Multi-gender) both trained from scratch and fine-tuned, also with gradient reversal and audio manipulation.}
\label{tab:results}
\end{table*}

% \begin{table*}[t]
% \small
% \centering
% \setlength{\tabcolsep}{10pt}
% \begin{tabular}{l|c|>{\centering}m{1.2cm} >% {\centering}m{1.2cm}||cc}
% \multirow{2}{*}{\textbf{Model}} & \multirow{2}{*}% {\textbf{BLEU}} & \multicolumn{4}{c}{\textbf{Gender % Accuracy}} \\
% \cline{3-6}
%  & & \multicolumn{1}{c}{\textbf{1F}} & \multicolumn{1}% {c||}{\textbf{1M}} & \multicolumn{1}{c}{\textbf{1F-% Tag M}} & \multicolumn{1}{c}{\textbf{1M-Tag F}} \\
%  \hline
% 
% \multicolumn{6}{c}{\textit{Trained from scratch}} \\
% \hline
% 
% \df{M}ulti\df{-}gender  & \textbf{27.2}  & % \textbf{84.0}  & 92.7  & 93.4  & \textbf{69.0}  \\
% \df{+ gradient reversal} & 24.9  & 70.9  & % \textbf{93.2}  & \textbf{94.1}  & 58.7  \\
% \df{+ audio manipulation (50\%)} & 26.2  & 79.6  & % 92.4  & 91.5  & 67.9  \\
% \df{+ audio manipulation (80\%)} & 25.7  & 81.7  & % 92.6  & 91.0  & 65.3  \\
% \hline
% \multicolumn{6}{c}{\textit{Finetuning}} \\
% \hline
% \df{S}pecialized  & \textbf{27.4}  & \textbf{73.3}  & % \textbf{92.5}   & \textbf{80.9}  & \textbf{56.1}   \\
% \df{M}ulti\df{-}gender   & 26.0  & 66.8  & 78.0  & % 64.6  & 47.1  \\
% + gradient reversal  & 26.7  & 60.7  & 85.9  & 77.0  % & 43.3  \\
% + gradient reversal weighted & 26.3  & 62.4  & 83.5  % & 77.7  & 45.9  \\
% \df{+ audio manipulation (50\%)} & 26.4  & 56.0  & % 82.6  & 60.7  & 33.9  \\
% \df{+ audio manipulation (80\%)} & 26.3  & 69.3  & % 81.1  & 69.0  & 47.7
% 
% \end{tabular}
% \end{table*}

\section{Experimental Settings}

%Our experiments have been carried out with
Our ST models are composed of
a Conformer \cite{gulati20_interspeech} encoder with 12 layers and a Transformer \cite{transformer} decoder with 6 layers. We used the Conformer implementation by \cite{papi2023reproducibility}, which does not contain bugs related to the presence of padding. The embedding size was 512, and the dropout was set to 0.1. We optimized label-smoothed cross entropy using Adam. The learning rate followed the Noam scheduler with 25,000 warmup updates and a maximum value of $2e^{-3}$. We train for 50,000 updates and average the last 7 checkpoints.

We train our models on MuST-C \cite{CATTONI2021101155}, an ST corpus built from TED data, 
for which is also available the annotation of the gender of the speaker \cite{gaido-etal-2020-breeding}.
% \df{which also includes annotations for the gender of the speakers.}
We extract 80 features with log mel-filterbank from the input audio and normalize them with cepstral mean and variance \cite{VIIKKI1998133}. The target text is encoded into subwords with 8,000 BPE merge rules \cite{di-gangi-etal-2020-target} learned on the training set.
We evaluate on the MuST-SHE benchmark \cite{bentivogli-etal-2020-gender}, which contains a section (``Category 1'') dedicated to assessing the gender assignment of words referring to the speaker.
%\df{through \textit{gender accuracy}, i.e. the percentage of words generated in the correct gender \cite{gaido-etal-2020-breeding}}.
We compute SacreBLEU\footnote{case:mixed$\vert$eff:no$\vert$tok:13a$\vert$smooth:exp$\vert$version:2.0.0} \cite{post-2018-call} on the whole MuST-SHE test set to evaluate the translation quality of our models and gender accuracy 
\cite{gaido-etal-2020-breeding} 
on the feminine and masculine sections of 
``Category 1''
to evaluate the ability of each model to correctly assign gender to words referring to the speaker.

% \df{To investigate the integration of gender metadata into the ST models, we employed % different training strategies. 
% Firstly, we trained multi-gender models in two ways: fine-tuning a base ST model on MuST-C % with each segment augmented with the gender tag, and training them from scratch.
% Secondly, for all the multi-gender models (both fine-tuned and trained from scratch), we % utilized the gradient reversal layer and the audio manipulation technique separately as % additional strategies.
% }

\paragraph{Gradient Reversal.}
The loss of the auxiliary speaker-classification task is summed to the loss on the decoder output scaling it by a $0.5$ factor.
For the gradient reversal layer, we tested both fixed values of $\lambda$ and controlling its value with $\gamma$. For fine-tunings, we set $\lambda=10$, so as to give similar weight to the gender classification loss and the cross-entropy loss for the translation. When training from scratch, instead, despite different attempts the training is unstable and diverges unless lambda is set to a fixed, small value, where its contribution is negligible. We report results for $\lambda=0.5$, which is the highest $\lambda$ value for which the loss on the validation set does not explode during training.

\paragraph{Audio Manipulation.}
In our experiments, we tested two values (0.5 and 0.8) for the hyperparameter $p$, which controls the probability of manipulating a speech segment. 
In the first case, 
%50\% of data are manipulated, so the correlation between gender tags (and target translations) and vocal traits is completely lost 
50\% of the data is manipulated, leading to a complete loss of correlation between gender tags and vocal traits
%(50\% of the samples with the F tag would show male voices and 50\% of the sample with the M tag would have female voices).
(50\% of the samples with the F tag would exhibit frequency characteristics typical of masculine voices, and 50\% of the samples with the M tag would have frequency characteristics typical of feminine voices).
In the second case, instead, the correlation between the gender tag and the vocal traits 
is 
%\df{becomes}
negative, 
to 
%\df{further}
counteract 
%even more 
the patterns learned by a gender-unaware ST model.
%This means that, in the first case, 50\% of data are manipulated, while, in the second case, 80\% of the segments are converted. 
% Since the manipulation probability is the same for segments uttered by both male and female speakers, the gender imbalance between in the training data is not mitigated.
In any case, as the training data is imbalanced (70\% of the samples are uttered by male speakers, and 30\% by female speakers), and the manipulation probability is the same for segments uttered by male and female speakers,  the gender imbalance in the training data is not mitigated.
%\df{Finally, as a comparison for our multi-gender models, we re-implemented the specialized models described in \cite{gaido-etal-2020-breeding}.}

\section{Results}

We investigate the performance of multi-gender models trained in two different ways: \textit{i)} fine-tuning a base, gender-unaware ST model, and \textit{ii)} training from scratch. 
In both cases, we study the effect of the introduction of the discriminator with gradient reversal and of the audio manipulation techniques.
% \dfff{For both approaches, we report the scores obtained when incorporating the discriminator with gradient reversal and the audio manipulation technique.}
% 
% \df{To investigate the integration of gender metadata into the ST models, we employed % different training strategies. 
% Firstly, we trained multi-gender models in two ways: fine-tuning a base ST model on MuST-C % with each segment augmented with the gender tag, and training them from scratch.
% Secondly, for all the multi-gender models (both fine-tuned and trained from scratch), we % utilized the gradient reversal layer and the audio manipulation technique separately as % additional strategies.}
Table \ref{tab:results} presents BLEU and gender accuracy scores 
(separately for segments spoken by female (1F) and male (1M) speakers)
for all the models, comparing them with the specialized models.
% the specialized models and the multi-gender models trained using various investigated strategies.
%BLEU is computed on the whole MuST-SHE benchmark, while
%gender accuracy is reported on Category 1 separately for segments spoken by female (1F) and male (1M) speakers.
To assess the inclusivity of our solution in cases where speakers exhibit vocal traits that do not conform with traditional gender perceptions, we also report gender accuracy for tests where the gender tag is inverted compared to the original audio segment (1F-Tag M and 1M-Tag F).
In these instances, the gender translation is expected to align with the gender tag, and we use the ``wrong'' reference of MuST-SHE, which swaps the speaker's references to the opposite gender.

\paragraph{Fine-tuning.}
The fine-tuned models from a base ST system consistently yield lower scores compared to the specialized systems. 
The simple multi-gender model performs 1.4 BLEU points worse than the specialized models in terms of overall translation quality. 
However, when audio manipulation and especially gradient reversal techniques are employed during fine-tuning, the performance gap is reduced by up to half.
Regarding gender accuracy, the multi-gender model achieves considerably lower scores than the specialized models, confirming previous findings from \cite{gaido-etal-2020-breeding}. 
This indicates that a fine-tuned multi-gender model struggles to accurately follow the gender tag for gender translation.
The accuracy gap is particularly high when the tag conflicts with vocal traits (-16.3 in 1F-Tag M, -9.0 in 1M-Tag F), where multi-gender models show below-chance accuracy for feminine forms, being below 50\%.
% 
% for masculine translations, where it is 14.5 points in category 1M, and 16.3 points in category 1F-Tag M. For feminine translations, the gap is 6.5 points in category 1F, and 9.0 points in category 1M-Tag F.
Both gradient reversal and audio manipulation techniques seem to further bias the model towards masculine forms. This likely indicates that the reduced ability to rely on the speakers' vocal traits is not compensated by the model looking at the gender tag, rather it strengthens its tendency to default to the most frequent masculine forms. 
% The gender tag alone does not prevent this bias, as the model struggles to follow it. When the gender tags are inverted, the same trend is observed, with higher scores for category 1F-Tag M.
The only technique that consistently improves both masculine and feminine translations compared to the simple fine-tuned multi-gender model is the introduction of audio manipulation with high probability (80\%). However, the gains (2.5 for 1F and 3.1 for 1M, and 4.4 for 1F-Tag M and 0.7 for 1M-Tag F) are limited and the gap with specialized models remains large. 
%However, even with these improvements, there are still considerable gaps with the specialized models.

%
\paragraph{Training from scratch.}
Unlike the fine-tuned models, the multi-gender models trained from scratch yield comparable or even higher results than the specialized models. This suggests that when an ST model is trained from scratch with gender tags, it learns to effectively follow them. Specifically, the simple multi-gender model trained from scratch achieves comparable translation quality to the specialized system (-0.2 BLEU) and significantly outperforms it in gender accuracy, with gains ranging from 0.2 (1M) to 12.5 (1F-Tag M).
% shows a BLEU score comparable to the specialized system (-0.2 points) and significant improvements in gender accuracy.
%The accuracy gains range from 0.2 points for category 1M, where the specialized models already perform well, to 12.5 points in category 1F-Tag M.
%
As in the fine-tuning case, neither gradient reversal nor audio manipulations increase the reliance of the model on the tag and the resulting models are more biased toward masculine forms. In fact, the multi-gender with gradient reversal reaches the highest accuracies in producing masculine forms (93.2 in 1M and 94.1 in 1F-Tag M), while suffering substantial drops in feminine accuracy. This effect is reduced when the weight of the F class is increased in the discriminator. In addition, in both cases the translation quality suffers a considerable drop. In the case of audio manipulation, the translation quality drop is lower (although still present), as well as the differences in terms of gender accuracy. They do not provide, though, any benefit compared to the simple multi-gender training.

% Regarding gender accuracy, the gradient reversal technique introduces bias towards the % masculine, with improvements of up to 0.7 points in category 1F-Tag M, but it decreases % feminine translation accuracy by up to 13.1 points. This indicates that the model struggles to % follow the gender tag when the gradient reversal technique is used, like in the fine-tuned % model. 
% On the other hand, audio manipulation leads to decreases in gender accuracy, with drops of up % to 3.7 points in category 1M-Tag F.

In summary, the results demonstrate that the previous finding about the low performance of multi-gender models is due to the adoption of a fine-tuning strategy. In this setting,
%the non-competitive multi-gender scores are primarily due to the simple fine-tuning approach,
%where 
the model cannot effectively override the reliance on speakers' vocal traits of the gender-unaware base ST model.
In addition, techniques aimed at avoiding the exploitation of speakers' vocal traits seem ineffective.
However, training the multi-gender model effectively solves the problem and the model is capable of following the indication given by the gender tag, outperforming even the specialized strategy by up to 12.9 gender accuracy (1M-Tag F).
% from scratch highlights the importance and usefulness of using the gender tag.
%
% Techniques for removing the gender attribute, such as gradient reversal, do not appear to be % effective and can sometimes conflict with the use of the gender tag. To enable the model to % prioritize the gender tag over acoustic information and achieve better results, further % investigation into alternative strategies is necessary.

% https://docs.google.com/spreadsheets/d/1WuD_JnHBDeCAuwp6OmCwwuEMId2okxfgxt7QZ979x_o/edit#gid=2132193396

\section{Conclusions}

In this paper, we studied the effect of different training strategies to build multi-gender ST models, i.e. models that are informed of the gender of the speaker by an explicit gender tag. 
%
%explored various training techniques to develop competitive ST models that utilize an explicit gender tag to translate the gender of speaker-dependent words accurately. 
Focusing on English-Italian translations, we demonstrated that the low accuracy of multi-gender models shown by previous work stems from the 
%
%identified that the low accuracy of multi-gender models stemmed from 
their initialization with gender-unaware ST system weights
%. This prevented them from 
and the inability of
effectively overriding the reliance on vocal cues during fine-tuning. 
On the other hand, training multi-gender models from scratch proved 
%to be more successful, 
to be an effective solution,
outperforming 
%other concurrent approaches 
the approach based on the creation of two gender-specialized models.
%in guiding the gender translation of speaker-dependent words.
%
As training from scratch is not always feasible,
%To address the reliance on acoustic information in multi-gender models,
we also experimented with two methods
to enhance the reliance on the gender tag in fine-tuned multi-gender models:
%a contrastive loss, 
penalizing the extraction of gender cues from speech input, and 
%also altered 
altering the vocal properties of the speakers in the training data to avoid the alignment between biological cues and gender tags and translations. 
While these solutions partially improved gender accuracy and overall translation quality in fine-tuned multi-gender models, they did not close the gap with specialized models.
%still did not completely overcome the issue of overriding the reliance on vocal cues during fine-tuning. 
Therefore, further research is needed in this direction.
%to develop more effective strategies.


\begin{acknowledgments}
  This work is part of the project ``Bias Mitigation and Gender Neutralization Techniques for Automatic Translation'', which is financially supported by an Amazon Research Award AWS AI grant. 
We acknowledge the support of the PNRR project FAIR -  Future AI Research (PE00000013),  under the NRRP MUR program funded by the NextGenerationEU.
\end{acknowledgments}

%%
%% Define the bibliography file to be used
\bibliography{sample-ceur}


\end{document}